\documentclass{article}
\usepackage{spconf,amsmath,epsfig}
\usepackage{amsmath,amsfonts,amssymb}
\usepackage{subcaption}
\usepackage{multirow}
\usepackage{booktabs}
\usepackage{xcolor}

\usepackage[normalem]{ulem}
\usepackage{url}
\usepackage{fancyhdr}

\title{Self Supervised Learning for Few Shot Hyperspectral Image Classification}
\name{Nassim Ait Ali Braham \textsuperscript{1}, Lichao Mou\textsuperscript{1, 2}, Jocelyn Chanussot \textsuperscript{3}, Julien Mairal \textsuperscript{3}, Xiao Xiang Zhu\textsuperscript{1,2}}
\address{\textsuperscript{1}German Aerospace Center (DLR), Remote Sensing Technology Institute (IMF), Wessling, Germany \\
\textsuperscript{2}Technical University of Munich (TUM), Data Science in Earth Observation (SiPEO), Munich, Germany \\
\textsuperscript{3}Inria, Univ. Grenoble Alpes, CNRS, Grenoble INP, LJK, Grenoble, France
}
\begin{document}
%
\maketitle
\begin{abstract}
    Deep learning has proven to be a very effective approach for Hyperspectral Image (HSI) classification. However, deep neural networks require large annotated datasets to generalize well. This limits the applicability of deep learning for HSI classification, where manually labelling thousands of pixels for every scene is impractical. In this paper, we propose to leverage Self Supervised Learning (SSL) for HSI classification. We show that by pre-training an encoder on unlabeled pixels using Barlow-Twins, a state-of-the-art SSL algorithm, we can obtain accurate models with a handful of labels. Experimental results demonstrate that this approach significantly outperforms vanilla supervised learning.
\end{abstract}

\fancypagestyle{copyright}{%
    \fancyhf{}\renewcommand{\headrulewidth}{0pt}
    \cfoot{\footnotesize%
        \textcopyright~2022 IEEE. Personal use of this
        material is permitted. Permission from IEEE must be obtained for all other
        uses, in any current or future media, including reprinting\slash republishing
        this material for advertising or promotional purposes, creating new collective
        works, for resale or redistribution to servers or lists, or reuse of any
        copyrighted component of this work in other works. 
    }
}
\thispagestyle{copyright}

\section{Introduction}
\label{sec:intro}
    Hyperspectral images can capture very rich information about the physical characteristics of objects in a scene. This unique property of hyperspectral data, coupled with an increasing availability of cost-effective sensors, and an improving spatial and spectral resolutions, has enabled many applications in agriculture, environmental monitoring, biomedical imaging, and many others. This has called for the development of many hyperspectral image analysis algorithms, especially for HSI classification.
    
    HSI classification 
    has received considerable attention from the remote sensing community. 
    Earlier approaches typically relied on a two-steps procedure: $(i)$ a feature extraction step; 
    $(ii)$ a shallow classifier such as SVM. Nowadays, deep learning has become the dominant approach for this problem thanks to the ability of neural networks to extract highly non-linear relationships from raw data 
    
    Unfortunately, training deep models following the supervised learning paradigm requires large amounts of well annotated samples to generalize and avoid overfitting. On the other hand, accurately labelling thousands of pixels in a hyperspectral image is very tedious, costly, and requires expert knowledge. This limitation can severely impede the applicability of deep learning on hyperspectral data. To mitigate this issue, a lot of efforts have been devoted to develop more label efficient methodologies, such as semi-supervised learning, meta-learning and weakly-supervised learning. These paradigms have been utilized on hyperspectral data as well \cite{mou2017unsupervised, wu2017semi, liu2018deep} showing promising results. Yet, the problem of label efficient HSI classification remains open.  
    
    In this paper, we propose to leverage recent progress in Self Supervised Learning (SSL) \cite{jing2020self} to enable accurate HSI classification with limited labels. Specifically, we show that state-of-the-art SSL algorithms from the computer vision literature can be used to pre-train models using unlabeled pixels in a scene. By doing so, one can quickly adapt such pre-trained models for classification with only few labels. 
    In addition, we demonstrate how inductive priors, such as the spatial regularity of a scene, can be easily incorporated into the pre-training stage through an appropriate pair-sampling strategy. Finally, we also analyze the impact of data augmentation on the performance of SSL on hyperspectral data. Our approach is compatible with all joint-embedding SSL methods and can be applied to pixel-level and patch-level HSI classification. The results we obtain show that the proposed pipeline is far superior to a supervised learning baseline.  
    
\vspace{-5mm}
\section{Proposed Approach}
\vspace{-3mm}
\label{sec:format}
    \subsection{Overview}
\begin{figure*}
            \centering
            \includegraphics[width=\textwidth]{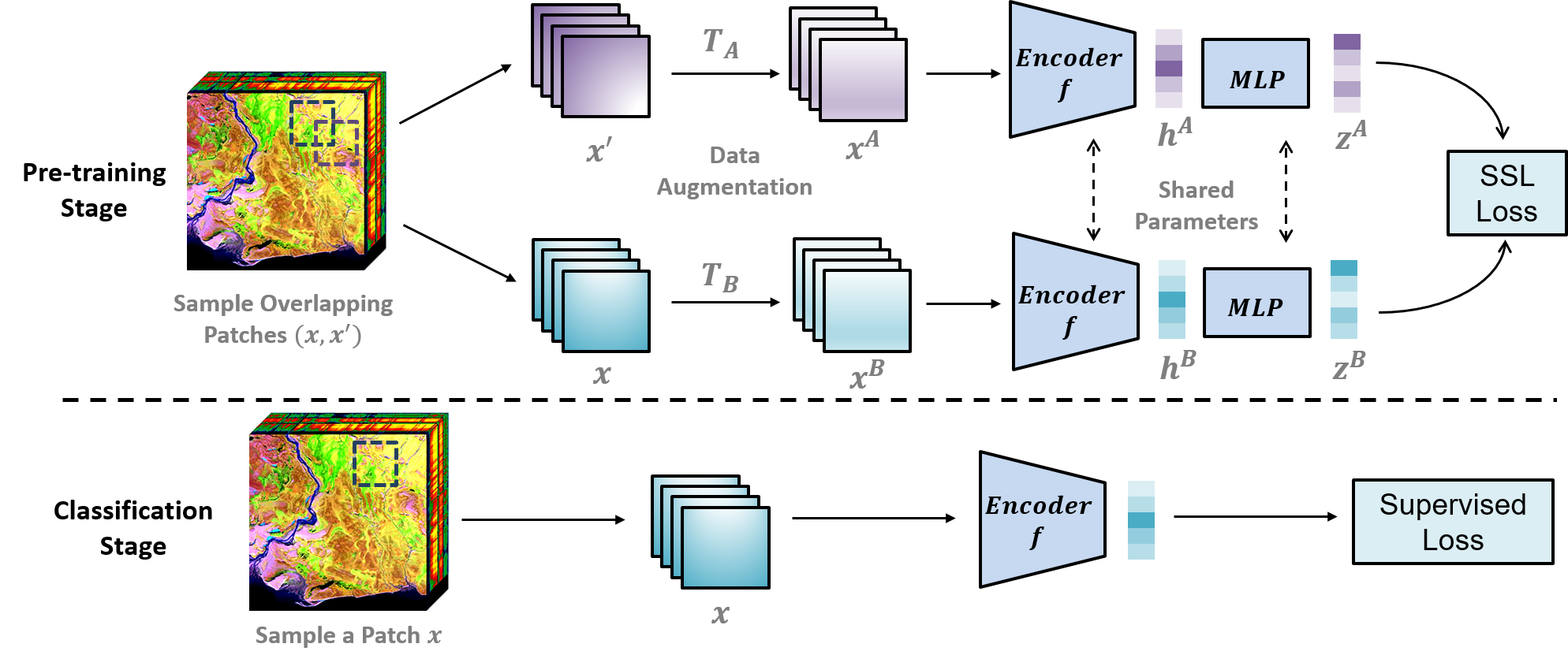}
            \caption{Overview of the proposed approach: pre-train using SSL and use the resulting encoder for HSI classification.}
            \label{fig:overview}
            \vspace{-1mm}
        \end{figure*}

        Let $X \in \mathbb{R}^{N \times M \times C}$ be a hyperspectral image. We denote by $x \in \mathbb{R}^{p \times p \times d}$ a hypercube of size $p \times p$ sampled from the scene (or a pixel when $p = 1$). The goal is to pre-train an encoder $f$ on unlabeled pixels using SSL, and use $f$ to train a classification model $g$ on the set of labeled pixels. The overall approach is depicted in Fig. \ref{fig:overview}.

        We follow a two-stages procedure consisting of a pre-training and a classification phase. It is a classical approach in unsupervised learning which has also been used for HSI classification \cite{mou2017unsupervised}. However, modern SSL methods are still under-studied on hyperspectral data, even though some recent studies for clustering and classification are emerging \cite{cai2021large, liu2020deep}.
    \subsection{Pre-training Stage}
            For pre-training, we treat all pixels as unlabeled and train an encoder $f$ using Barlow-Twins \cite{zbontar2021barlow}, which we describe below. 
            
            Barlow-Twins is an SSL algorithm based on redundancy minimization.  
            Specifically, from an input $x$, two views $x^A = T_A(x)$ and $x^B=T_B(x)$ are stochastically generated using two sets of augmentations $T_A, T_B$. In this work, we also leverage spatial information to create the views by sampling partially overlapping patch pairs (see Section \ref{sec:views}). The views are fed into a shared encoder $f$ to obtain their representations $h^A = f(x^A)$ and $h^B = f(x^B)$. Then, an additional Multi-Layer Percepteron (MLP) head is applied to get $z^A=MLP(h^A)$ and $z^B=MLP(h^B)$. Since $x^A$ and $x^B$ are augmented views, their representations should be similar. Moreover, following the redundancy minimization principle, the dimensions of $z^A$ and $z^B$ should be decorrelated. This can be enforced on a batch-level by optimizing the following loss: 
            \vspace{-0.5mm}
            \begin{equation}
                \mathcal{L} = \underbrace{\sum_{i}\left(1-\mathcal{C}_{i i}\right)^{2}}_{\text {invariance to augmentation}}+\ \lambda \underbrace{\sum_{i} \sum_{j \neq i} \mathcal{C}_{i j}^{2}}_{\text {redundancy reduction }}
            \end{equation}
        \vspace{-1.5mm}
            
        \noindent where $\lambda \in \mathbb{R^+}$ is a trade-off parameter and $C$ is the cross-correlation matrix computed between the outputs $Z^A$ and $Z^B$ of the network branches along the batch dimension:
        \vspace{-1mm}
        \begin{equation}
            \mathcal{C}_{i j} = \frac{\sum_{b} z_{b, i}^{A} z_{b, j}^{B}}{\sqrt{\sum_{b}\left(z_{b, i}^{A}\right)^{2}} \sqrt{\sum_{b}\left(z_{b, j}^{B}\right)^{2}}}
        \end{equation}
    
    \vspace{-3mm}
    \subsection{Classification Stage}
            The pre-training step produces an encoder $f$ which has learnt useful representations for the pixels in the scene. The goal of the second step is to leverage this pre-trained model to construct a classifier $g$. For this purpose, we use two common approaches: linear classification and finetuning. 
            
            \begin{enumerate}
                \item \textbf{Linear Classification: } the linear protocol constructs a linear classifier $l$ on top of the representations produced by $f$. The classifier is $g = l \circ f$, where $f$ is kept frozen. 
                \item \textbf{Finetuning: } the finetuning protocol uses the encoder's weights as an initialization to train the classifier $g = l \circ f$. All the parameters of $f$ are trainable. Therefore, special care must be taken to avoid overfitting or moving away too quickly from the initial pre-trained model. 
            \end{enumerate}
    \vspace{-4mm}
    \subsection{Views generation}
    \label{sec:views}
        SSL algorithms heavily depend on the view-generation strategy, which is traditionally done by stochastically augmenting the same input. 
        In this work, we leverage spatial information combined with data augmentation to generate the views.  
        \vspace{-1.5mm}
        \subsubsection{Pair Sampling}
            Since patches come from a unique scene, we can exploit spatial locality to sample pairs before data augmentation. Specifically, instead of augmenting the same hypercube $x$, we take $x$ and an overlapping patch $x'$ as a pair. By ensuring a reasonable overlap between $x$ and $x'$ (e.g., $50\%$ at least), we can safely assume that they should have similar representations. 
            
            A similar idea can be applied at pixel-level. For example, given a pixel $x$, we propose to randomly select another pixel $x'$ from a pre-defined neighborhood and use $(x, x')$ as a pair.  
            
            Such sampling strategies introduce a spatial regularity prior into pre-training. Moreover, they provide a supervision signal without data augmentation, which reduces the need for aggressive transformations as commonly done in SSL. Similar ideas have been used for SSL on satellite imagery \cite{jean2019tile2vec}.

        \begin{figure}[htb]
            \centering
            \includegraphics[width=0.45\textwidth]{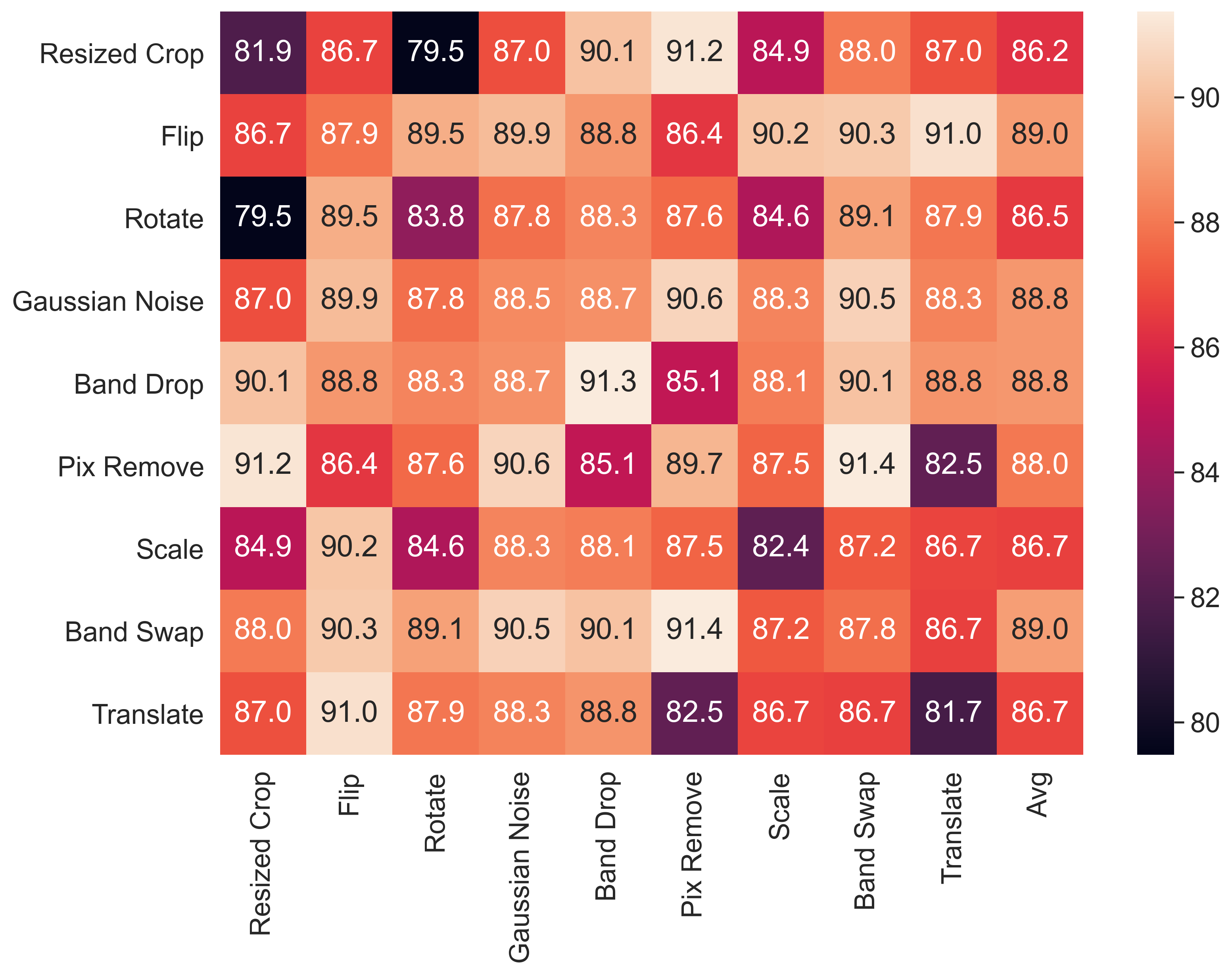}
            \caption{Impact of data augmentation on SSL for PaviaU.}
            \label{fig:augmatrix}
            \vspace{-2.1mm}
        \end{figure}
        
        \subsubsection{Data Augmentation}
            We consider two classes of transforms in this work: spatial and spectral augmentations. Spatial augmentations require a sufficiently large spatial context, but they preserve spectral information. Spectral transform on the other hand can be used for individual pixels, but may distort the semantics of the data. Below we list the augmentations we experiment with.
            \begin{enumerate}
                \item \textbf{Spatial Transforms:} random flipping (horizontal or vertical), random rotation (by multiples of $90$ degrees), random resized crop. These are identical to classical transforms on images and implemented in PyTorch.
                \vspace{-0.8mm}
                \item \textbf{Spectral Transforms:} scaling, Gaussian noise, random band dropping, random pixel removal, random band swapping (adjacent bands) and random translation (adding a random bias to the spectrum). 
            \end{enumerate}
            \vspace{-2mm}

\section{Experimental Setting}
\label{sec:pagestyle}
    \subsection{Data}
        We evaluate the approach on Pavia and Houston University datasets. Pavia University is a $610 \times 340$ pixels scene with 9 classes and 103 bands ranging from 0.43 to 0.86 µm. Houston University is a 349 × 1905  with 144 bands ranging from 0.36 to 1.04 µm and 15 classes. All bands are used in our experiments without any pre-processing (besides normalization).
            
        For both datasets, we use multiple training maps. Specifically, we consider a few-shot setting where we randomly select $K \in [5, 10]$ labeled pixels per class. In addition, we consider an abundant labels scenario using the training maps provided in DASE\footnote{\url{http://dase.grss-ieee.org/}}, which are much more challenging than random sampling. 
        
    \subsection{Models and Metrics}
        We consider simple models for HSI classification. Specifically, in the patch-level setting, we rely on a custom 2D Convolutional Neural Network (CNN) with 2 residual blocks. The patch-size we use is $9 \times 9$. For the 1D case, we use a similar 1D CNN architecture with 2 residual blocks. 
        
        In all experiments, Barlow-Twins pre-training is ran for 100 epochs with LARS optimizer. Classification models are trained for 100 epochs using stochastic gradient descent with a cross-entropy loss. We report the Overall Accuracy (OA) and the Kappa coefficient ($\kappa$) in the experiments and compare against a supervised baseline using the same networks.

\section{Experimental Results}

    \subsection{Impact of Data Augmentation}
        We analyze the impact of data augmentation on the overall accuracy of Barlow-Twins on PaviaU with a linear classifier using the training map from DASE. To do so, we consider up to two augmentations at a time. We run the pre-training and classification phase for every pair of transformations listed in Section \ref{sec:views}. The transforms are applied in both branches in a symmetric manner with a fixed order and probability $p = 0.75$. Results are presented as a symmetric matrix in Fig. \ref{fig:augmatrix}. 
        
        We observe that the transforms which work best on average are random flipping, Gaussian noise, random band dropping and random band swapping. Moreover, a part from a few exceptions, the diagonal values of the matrix (i.e., when a single transform is used) are usually lower than the off-diagonals (i.e., transformation pairs). Additionally, when discarding all transformations, we get $79.4\%$ of accuracy, which confirms the need for data augmentation. 
        Yet, accumulating many transforms can be problematic with hyperspectral images since, contrarily to RGB data, the relevant information lies in the spectrum of every pixel. 
        
    \begin{table}[htbp]
        \setlength{\tabcolsep}{4pt}
        \centering
        \begin{tabular}{|p{1.5cm}|p{0.5cm}|c|c|c|c|c|c|}
            \hline
            \multicolumn{2}{|c|}{} & K=5 & K=6 & K=7 & K=8 & K=9 & K=10  \\ \hline
            \multirow{2}{*}{Supervised} & OA & 61.2 & 67.3 & 59.6 & 69.4 & 76.8 & 74.7  \\
             & $\kappa$ & 52.2 & 58.7 & 50.3 & 62.3 & 70.0 & 67.4 \\ 
            \hline
            \multirow{2}{*}{BT + Lin} & OA & 80.6 & 82.9 & 86.7 & 84.9 & 87.2 & 90.3 \\ 
            & $\kappa$ & 75.3 & 78.4 & 82.8 & 80.8 & 83.6 & 87.4 \\
            \hline
            \multirow{2}{*}{BT + Tune} & OA & 79.2 & 82.3 & 87.0 & 83.5 & 87.7 & 89.9  \\ 
            & $\kappa$ & 73.2 & 77.6 & 83.2 & 79.3 & 84.1 & 86.9 \\
           \hline
        \end{tabular}
        \caption{Classification results on Pavia University with limited labels using 2D CNNs.}
        \label{results:pavia2d}
        \vspace{-4mm}
    \end{table}
    
    \begin{table}[htbp]
        \setlength{\tabcolsep}{4pt}
        \centering
        \begin{tabular}{|p{1.5cm}|p{0.5cm}|c|c|c|c|c|c|}
            \hline
            \multicolumn{2}{|c|}{} & K=5 & K=6 & K=7 & K=8 & K=9 & K=10  \\
            \hline
            \multirow{2}{*}{Supervised} & OA & 60.0 & 65.9 & 64.9 & 68.7 & 71.4 & 75.0 \\
             & $\kappa$ & 56.9 & 63.2 & 62.2 & 66.3 & 69.2 & 73.1 \\ \hline
            \multirow{2}{*}{BT + Lin} & OA & 74.4 & 73.3 & 77.9 & 74.1 & 77.0 &	80.5 \\ 
            & $\kappa$ & 72.5 & 71.2 & 76.2 & 72.2 & 75.2 & 79.0 \\
            \hline
            \multirow{2}{*}{BT + Tune} & OA & 72.9 & 72.5 &	76.2 & 74.2 & 76.1 & 82.0 \\ 
            & $\kappa$ & 70.7 & 70.4 & 74.4 & 72.3 & 74.3 & 80.7 \\
           \hline
        \end{tabular}
        \caption{Classification results on Houston University with limited labels using 2D CNNs.}
        \label{results:houston2d}
        \vspace{-4mm}
    \end{table}
    
    \subsection{2D Results}
        We run the proposed SSL-based approach on PaviaU and Houston with a varying number of samples per class using a 2D CNN. For these experiments, we use flipping, Gaussian noise, random band dropping and random band swapping for data augmentation. 
        Results are given in Table \ref{results:pavia2d} and Table \ref{results:houston2d}, where BT denotes Barlow-Twins. In every setting, SSL outperforms vanilla supervised learning by significant margins and proves to be more label efficient. Furthermore, the benefit of using SSL is also visible even when labels are abundant, as shown in Table \ref{results:paviahoustonfull}. Moreover, it is interesting to observe that the linear protocol (BT + Lin) and the finetuning protocol (BT + Tune) lead to comparable results, even-though most of the network parameters in the linear case are trained in an unsupervised way. This shows that the model has learnt useful features during the pre-training phase. In addition, the linear protocol is also less prone to overfitting. 
        
        \begin{table}[htbp]
        \centering
        \begin{tabular}{|c|c|c|c|c|}
            \hline
            \multirow{2}{*}{} & \multicolumn{2}{|c|}{PaviaU} & \multicolumn{2}{|c|}{Houston} \\
            \hline
            & OA & $\kappa$ & OA & $\kappa$ \\
            \hline
            Supervised & 88.4 & 84.8 & 77.6 & 75.8 \\
            \hline
            BT + Lin & 88.9 & 85.1 & 78.4 & 76.6 \\
            \hline
            BT + Tune & 88.5 & 84.7 & 81.0 & 79.4 \\
            \hline
            
        \end{tabular}
        \caption{Classification results on Pavia and Houston University with large training maps using 2D CNNs.}
        \label{results:paviahoustonfull}
       
    \end{table}
    \vspace{-5mm}
    \subsection{1D Results}
        We train a BT + Lin classifier on PaviaU using 1D CNN. We generate the pairs using neighboring pixels (in a $5 \times 5$ neighborhood). We do not apply any data augmentation because preliminary experiments suggest that it does not always help with 1D CNNs. The results are shown in Table \ref{results:pavia1d}. Once again, we observe that SSL outperforms supervised learning when the labels are limited. It is also the case on the full split, where supervised learning reaches $80.9\%$ of accuracy compared to $86.2\%$ for BT + Lin.

    \begin{table}[htbp]
        \setlength{\tabcolsep}{4pt}
        \centering
        \begin{tabular}{|p{1.5cm}|p{0.5cm}|c|c|c|c|c|c|}
    
            \hline
            \multicolumn{2}{|c|}{} & K=5 & K=6 & K=7 & K=8 & K=9 & K=10  \\
            \hline
            \multirow{2}{*}{Supervised} & OA & 43.5 & 64.6 & 57.6 &	58.2 &	65.8 & 65.8 \\
             & $\kappa$ & 36.9 & 55.3 & 50.1 & 50.4 & 57.1 & 56.3 \\ \hline
            \multirow{2}{*}{BT + Lin} & OA & 70.0 & 77.6 & 86.8 &	83.5 & 85.8 & 88.0 \\ 
            & $\kappa$ & 63.1 & 71.9 & 82.8 & 79.0 & 81.8 & 84.5 \\
            \hline
        \end{tabular}
        \caption{Classification results on Pavia University with limited labels using 1D CNNs.}
        \label{results:pavia1d}
        \vspace{-4mm}
    \end{table}

\section{Conclusion}
\label{sec:conclusion}
In this paper, we have used Barlow-Twins, a state-of-the-art SSL algorithm, for few shot HSI classification. We have proposed pair sampling strategies for patches and pixels, and have investigated several data augmentations and their impact on performance. Our experimental results suggest that SSL outperforms plain supervised learning and improves label efficiency. Future work will investigate more data driven view generation strategies for SSL on hyperspectral data.  
\label{sec:ref}
\vspace{-2mm}
\bibliographystyle{IEEEbib}
\bibliography{strings,refs}

\end{document}